# Incremental Validation of Automated Driving Functions using Generic Volumes in Micro-Operational Design Domains

Steffen Schäfer[1], Martin Cichon
Karlsruhe Institute of Technology, KIT
Institute of Vehicle Systems Technology, FAST
Rintheimer Querallee 2, Bld. 70.04
76131 Karlsruhe, Germany
[1]s.schaefer@kit.edu (Corresponding Author)

*Abstract*— The validation of highly automated, perception-based driving systems must ensure that they function correctly under the full range of real-world conditions. Scenario-based testing is a prominent approach to addressing this challenge, as it involves the systematic simulation of objects and environments. Operational Design Domains (ODDs) are usually described using a taxonomy of qualitative designations for individual objects. However, the process of transitioning from taxonomy to concrete test cases remains unstructured, and completeness is theoretical. This paper introduces a structured method of subdividing the ODD into manageable sections, termed micro-ODDs (µODDs), and deriving test cases with abstract object representations. This concept is demonstrated using a one-dimensional, laterally guided manoeuvre involving a shunting locomotive within a constrained ODD. In this example, µODDs are defined and refined into narrow taxonomies that enable test case generation. Obstacles are represented as generic cubes of varying sizes, providing a simplified yet robust means of evaluating perception performance. A series of tests were conducted in a closed-loop, co-simulated virtual environment featuring photorealistic rendering and simulated LiDAR, GNSS and camera sensors. The results demonstrate how edge cases in obstacle detection can be systematically explored and how perception quality can be evaluated based on observed vehicle behaviour, using crash versus safe stop as the outcome metrics. These findings support the development of a standardised framework for safety argumentation and offer a practical step towards the validation and authorisation of automated driving functions.

## I. Introduction

The impact of automating railway operations is widely discussed and can even be observed in service for instance at sealed off systems such as metros[1], [2], [3]. The automating approaches vary from vehicle sided environment perception and control systems to land sided surveillance and remote-controlled solutions. As the rail network is very extensive and land-based automation is therefore expensive, most automated driving systems use a dedicated sensor system on the vehicle. In previous work, the authors have made significant progress in validating these automated systems, but a common legislative process has not yet been established [4]. In this context, a photorealistic simulation environment, modelled on the Munich North digital shunting yard, is introduced for virtual closed-loop laboratory testing [5]. The aim of this paper is to establish a step-by-step approach that can be used to systematically reduce the number of tests required to determine the limits of safe operation. Therefore, according to an established method, the complex ecosystem of the automated rail vehicle is defined as the operational design domain (ODD) and broken down for a specific use case. Further simplifications result in a problem description that is both understandable and treatable at the µODD level [6].

## II. Micro Operational Design Domain

In previous studies, the authors published a framework for defining an ODD for automated train operations, based on a dedicated taxonomy [7]. The larger the ecosystem and its eventualities are depicted, the more extensive and hence more unspecific this description becomes. For a wide and unspecific ODD, the number of test cases attempting to validate safe operation is uncountable. A promising approach to designing a straightforward bottom-up verification strategy is to formulate a mosaic of µODDs instead [8]. According to the

*"structure ODD taxonomy for railway systems"* [7]

the pillars are broken down as follows.

*A. Scenery*

Firstly, the area of operation is defined. For the sake of simplicity, the regular operation of a shunting locomotive can be limited to the processes in the arrival track group [4]. Following Open Railway Maps, the Munich track section is structured as schematically shown in Figure 1.

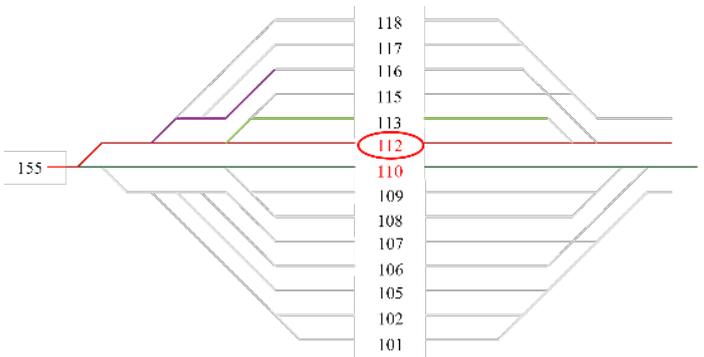

*Figure 1: Schematic layout of the Munich North arrival track section, red tracks 110 and 112 are comparable on their complexity*

Tracks 110 and 112 offer a good starting point for the design of an elementary μODD that nevertheless maps the minimum requirements of the shunting system. These tracks are comparable in complexity, combining representation of the track system with simplicity. Track 112 is chosen for this analysis, as the switch group is more densely packed here, the track straight is reached earlier and the total section is therefore shorter. This shunting yard is not equipped with a control system; vehicles are manoeuvred on sight. All drivable tracks and junctions must be in good condition, with a standard 1,435 mm gauge.

The area of the marshalling yard consists of lamp poles and some trees. There is no other vegetation or notable structures in the area. At the end of Track 155, facing the track harp, there is a light signal (red and white). The μODD considers switches to be passed to the left and right, respectively.

So far, the scenery does not include any other instances. Any temporary obstacles or objects in the area are not considered here, but are grouped with the dynamic elements under Section *D*.

*B. Environmental Conditions*

To minimise disturbance, sunny weather with a clear sky and clean air is chosen, with little wind and no obstacles blowing around. The time of day, and therefore the position of the sun, is between 10 a.m. and 2 p.m., which mainly affects the shadows cast, as well as the level of illumination. The environmental conditions are selected so that connectivity is affected as little as possible.

*C. Opeartional Conditions*

Building on the authors' previous work, regular freight train shunting operations can be clustered into a finite number of use cases [9]. The use case involving the least complex movement of the ego vehicle is 'Moving'. More challenging use cases involve physical interaction with other vehicles, such as attaching or pressing up. However, as a low-risk starting point, the 'Moving' use case and the associated driving path observation for crash avoidance are addressed within the first μODD. In this scenario, the vehicle is already in service and is waiting on the dead-end track (No. 155, Figure 1) for the next shunting task:

- Moving
- From track 155
- moving to track 112
- stop there.

*D. Dynamic Elements*

This section should be divided into regular services and anomalies, such as people, animals, or random obstacles within the driving gauge. As this use case does not consider the detection of rail vehicles for subsequent attachment, only anomalies are considered here. According to the literature, ODDs often involve explicit instances such as rabbits or other wildlife and objects [10, 11].

However, obstacle classification based on machine learning algorithms is not yet ready for licensing [12, 13]. Therefore, simple and fast driving decisions in response to obstacles depend on their presence or absence (quantity), rather than their classification (quality).

Following this argumentation based on quantity, the dynamic elements (DEs) considered within this μODD are represented by the simplest geometric body: a cube (see Figure 2). The right-hand side of the illustration shows how the cube can be placed at different points along the path.

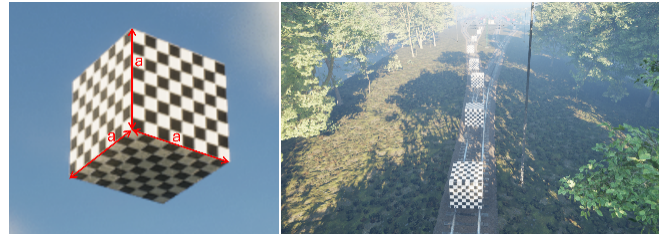

*Figure 2: Generic body. Cube alone with edge length a (left), installed in the virtual driving gauge in different distances (right)*

The second relevant parameter describing the DE, besides its position, is the cube's dimensions. Here, the obstacle size has been chosen to correspond to edge lengths between 500 mm and 3,000 mm. A value of $a = 500$ mm refers to the 'smallest object' that still needs to be recognised, as defined in references [14], [15] and [16], while a value of $a = 3,000$ mm roughly corresponds to the dimensions of a freight wagon.

### III. VIRTUAL CLOSED LOOP TEST SETUP

A virtual environment is used for test case setup and execution. The test cases are constructed for successive scenarios derived from the μODD. As explained in earlier publications by the authors, the test setup follows the closed-loop architecture [4] shown in Figure 3.

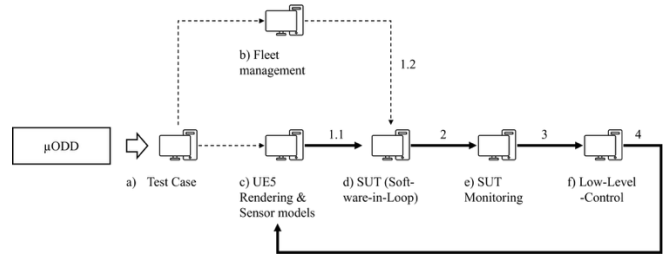

*Figure 3: Co-simulated Virtual Closed Loop Test Setup*

Building on the μODD approach described above, the complexity of the ODD is broken down into more manageable sub-areas. Concrete test cases are then developed based on this limited scope.

*A. Congrete Test Cases*

Research in ODD and taxonomy does not involve formulating concrete scenarios. To address this, a gradual approach has been introduced, whereby the entire route is divided into successive sections. This segmentation is based on changes to the driving gauge conditions. To this end, a curved x-axis that follows the ego track topology (here, track 112) is introduced (see Figure 4). The origin of the coordinate system is set to the dead-end track 155 and the y-axis points left, following the vehicle coordinate system [17]. The challenges for the automated system along the track differ and can even be grouped together. The first step to effective operation is proper signalling classification. The following scenarios feature variations in track topology and more complex scenery. The occurrence of each new element indicates a new scenario.

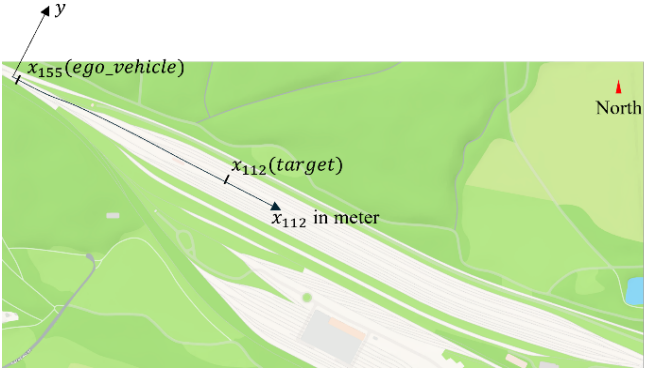

*Figure 4: Map of the Munich North shunting yard arrival tracks. Including the highlighted track from 155 into 112 introducing a curved x-axis description along the ego track*

Based on this, a descriptive visual language for test scenarios is introduced that depicts the ego track in a rectangular coordinate system (Figure 5). Following the x-axis from the origin — i.e. following the track from the dead end through the track switch harp into track 112 — reveals the local conditions that lead to scenario differentiation.

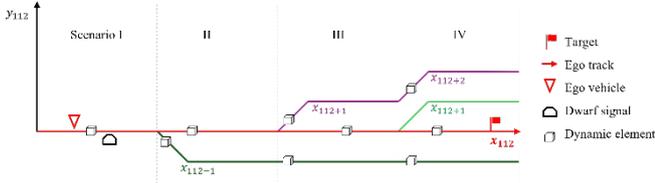

*Figure 5: Visual Scenario Description Language, introducing a projection of the ego track $x_i$ to a straight x-axis*

The start (Ego vehicle) and end (Target) position of the locomotive are set according to section II.C and are valid for each scenario:

*1) Scenario I:* Track 155 can be considered a reasonable approximation of a straight track. The horizon of the segment is set right behind the first dwarf signal (DS), indicating whether the driving gauge is free or locked behind. The pass/fail criteria are:

- Classify signal and stop if $DS = red$
- If $DS$ = *white:*
- Accelerate if $DE \notin x_{112}$
- Safe stop in front of $DE \in x_{112}$ (if existant)

*2) Scenario II:* At the segment boundary, the track harp begins. Track 110 ($x_{112-1}$) leaves the ego track to the right. The pass/ fail criteria are:

- Pass through if $DE \notin x_{112}$
- Pass through if $DE \in x_{112-1}(boundary\ sign)$
- Safe stop in front of DE (if $DE \in x_{112}$)

*3) Scenario III:* The topology becomes more complex as more tracks are added. Here track 113 ($x_{112+1}$) leaves the ego track to the left. The pass/fail criteria are similar to those described in the previous scenario:

- Pass through if $DE \notin x_{112}$
- Pass through if $DE \in x_{112+1}(boundary\ sign)$
- Pass through if $DE \in x_{112-1}$
- Safe stop in front of DE (if $DE \in x_{112}$)

*4) Scenario IV:* The track topology is even bigger now, and the target track is reached after passing the third switch point. The pass/fail criteria are similar to those described in the previous scenario:

- Drive in if $DE \notin x_{112}$
- Drive in if $DE \in x_{112+1}(boundary\ sign)$
- Drive in if $DE \in x_{112+2}$
- Drive in if $DE \in x_{112-1}$
- Safe stop in front of DE (if $DE \in x_{112}$)
- Stop in a safe distance after the switch $x_{112}(target)$

### B. Fleet Management

The control centre (CC) is responsible for initiating communication between the loco (initiated automated driving) and the infrastructure. The information sent to the vehicle (1.2) is formulated according to the order to be completed. In this case, the starting track (No. 155), the target track number and the task ID (2 [18]) for driving are transmitted to the system under test (d, SUT) via UDP.

### C. Virtual Rendering

The scenario has been set up and rendered using the Unreal Engine 5® (c). This represents the arrival tracks of the Munich North shunting yard including a virtual model of a locomotive. The vehicle model carries sensor models for LiDAR, RBG camera and GPS, which emulate the respective sensor data (1.1). This data is fed to the system under test (d, SUT) via UDP data streams.

### D. System Under Test

The SUT is a ROS2-Huble based software stack called KaITrain [18]. ROS2 communicates via the publisher-subscriber principle and topics (2) which can be monitored for human-machine interfacing (e) and prompted to the vehicle interface (3). KaITrain is designed for perception-based on sight driving of rail vehicles, which aligns with the overall ODD described in this work. For obstacle detection, a simple yet effective mechanism using the LiDAR point Cloud is employed. This basic principle does not involve classification, but rather monitoring to ensure the clearance gauge is free of obstacles. Therefore, the fast Euclidean clustering algorithm by Masashi Mizuno [19] has been embedded. This is a fast implementation of the well-known Euclidean clustering method [20]. Each detected cluster is treated as an object, if it exceeds a given threshold for the number of points to account for any given noise or incorrectly projected points. Checking the clearance gauge for any intersecting clusters reveals whether there are any objects within the driving path prompting the automated driving stack to make a decision. For now, this decision is limited by the ODD and its corresponding µODD, which determine whether to stay at speed or to slow down to a standstill. As the entire KaITrain detection pipeline is built modular, it is possibe to add filtering before clustering or classification after clustering is provided to aid decision-making in more complex scenarios in the future. However, the driving decision is forwarded to a vehicle model via UDP.

## E. Low-Level Control

Traditional automated driving stacks based on perception use a sensor unit to perceive the environment ('sense'), and are further structured for high-level decision-making ('plan') and low-level control of vehicle actuators, motors, and auxiliaries ('act') [21]. Within this virtual loop, both the low-level control and vehicle kinematics are simulated using a MATLAB Simulink model. To this end, the model receives the target velocity and the remaining distance to achieve the velocity delta, i.e. to slow down or accelerate, via UDP (3). The vehicle model then calculates the current actual speed and forwards it to the rendering environment (c) via UDP (4), where the speed information is converted directly into the movement of the ego locomotive.

## IV. RESULTS

For each of scenarios I–IV, the test cases were carried out in six increments. The edge length of the DM increased by 0.5 m at a time, from 0 m to 3 m.

TABLE I. TEST SCENARIO 1, OBSTACLE DISTANCE = 10 METERS, SIGNAL DISTANCE = 12 METERS

| Track 112 | Results Scenario I | | |
|---|---|---|---|
| | *Obstacle detected* | *Signal setting detected* | *Test Case Succeeded* |
| a = 0.0 m | - | ✓ | ✓ |
| a = 0.5 m | ✓ | ✗ | ✗ |
| a = 1.0 m | ✓ | ✗ | ✗ |
| a = 1.5 m | ✓ | ✗ | ✗ |
| a = 2.0 m | ✓ ✗ | ✗ | ✗ |
| a = 2.5 m | ✗ | ✗ | ✗ |
| a = 3.0 m | ✗ | ✗ | ✗ |

TABLE II. TEST SCENARIO 2, $DE \notin x_{112}, DE \in x_{112-1}(boundary\ sign)$

| Track 112 | Results Scenario II | | |
|---|---|---|---|
| | *Obstacle detected* | *Signal detected* | *Test Case Succeeded* |
| a = 0.0 m | - | ✓ | ✓ |
| a = 0.5 m | ✓ | ✓ | ✗ |
| a = 1.0 m | ✓ | ✓ | ✗ |
| a = 1.5 m | ✓ | ✗ | ✗ |
| a = 2.0 m | ✓ | ✗ | ✗ |
| a = 2.5 m | ✓ | ✗ | ✗ |
| a = 3.0 m | ✗ | ✗ | ✗ |

TABLE III. TEST SCENARIO 2, $DE \in x_{112}, DE \notin x_{112-1}(boundary\ sign)$

| Track 112 | Results Scenario II | | |
|---|---|---|---|
| | *Obstacle detected* | *Signal detected* | *Test Case Succeeded* |
| a = 0.0 m | - | ✓ | ✓ |
| a = 0.5 m | ✓ | ✓ | ✓ |
| a = 1.0 m | ✓ | ✓ | ✓ |
| a = 1.5 m | ✓ | ✓ | ✓ |
| a = 2.0 m | ✓ ✗ | ✓ | ✗ |
| a = 2.5 m | ✗ | ✓ | ✗ |
| a = 3.0 m | ✗ | ✓ | ✗ |

TABLE IV. TEST SCENARIO 4, $DE \in x_{112}, DE \notin \{x_{112-1}(boundary\ sign), x_{112+1}, x_{112+2}\}$

| Track 112 | Results Scenario II | | |
|---|---|---|---|
| | *Obstacle detected* | *Signal detected* | *Test Case Succeeded* |
| a = 0.0 m | - | ✓ | ✗ |
| a = 0.5 m | ✓ | ✓ | ✓ |
| a = 1.0 m | ✓ | ✓ | ✓ |
| a = 1.5 m | ✓ | ✓ | ✓ |
| a = 2.0 m | ✓ ✗ | ✓ | ✗ |
| a = 2.5 m | ✗ | ✓ | ✗ |
| a = 3.0 m | ✗ | ✓ | ✗ |

The results for Scenario 3 are comparable to those for Scenario 2 in terms of complexity. Eight scenario variations (with or without elements) are derived from the scenario descriptions, each of which is tested in seven test cases (a = 0 m – a = 3.0 m) across 56 test runs. Only the results relevant to describing the marginal cases are presented here; these can be used to improve the SUT.

## V. INTERPRETATION AND DISCUSSION

Starting with the results in Table 1, test case 1 (a = 0.0 m) demonstrates the functionality of the virtual closed-loop test bench. Further cases demonstrate that clustering works for small objects with an edge length of up to almost 2.0 m; however, detection begins to flicker when the DE size is incrementally increased. Larger objects are not clustered when the default parameters of the KaITrain clustering algorithms are used. Another interesting issue is that the KaITrain system's *observer_info* message incorrectly documents the signalling setting, even though the signal is correctly detected and published within the signal detector node.

Table 2 shows the results of the test runs where the neighbouring track is occupied from the point at which the boundary sign is reached. Similar results were obtained for test scenario 3. As illustrated in Figure 6, the edge of the obstacle is detected and interpreted as having been intersected by the clearance gauge; therefore, the vehicle is prompted to stop. However, the object perimeter does not actually cut the clearance gauge. This error can be attributed to incorrect mapping. The KaITrain system is based on an extremely simple digital map that provides only 27 support points for track 112 over the entire 1.5 km route up to the hump. Consequently, the arches are not reproduced in detail. However, when the clearance gauge is projected onto this map to check for collisions, the deviation from the actual track position causes a failure, resulting in the test run being cancelled. This effect is reduced the straighter the track section in front of the vehicle is.

Table 3 shows stable object detection and test case handling for all objects <2.0 m.

During these test runs, the signal is correctly detected and handled.

During the test cases derived from scenario IV, another error emerged. In the first test case (a = 0 m), the vehicle moves forward correctly into track 112 in response to the signal change from red to white. However, without an object in the path, the vehicle continues to travel and does not stop in track 112 as intended. This is due to this function not being implemented in the KaITrain system. The system is designed solely to recognise obstacles so that it can stop in front of them or approach them and never checks its own position with regard to a local target. As in previous tests, those involving objects <2.0 m are successful.

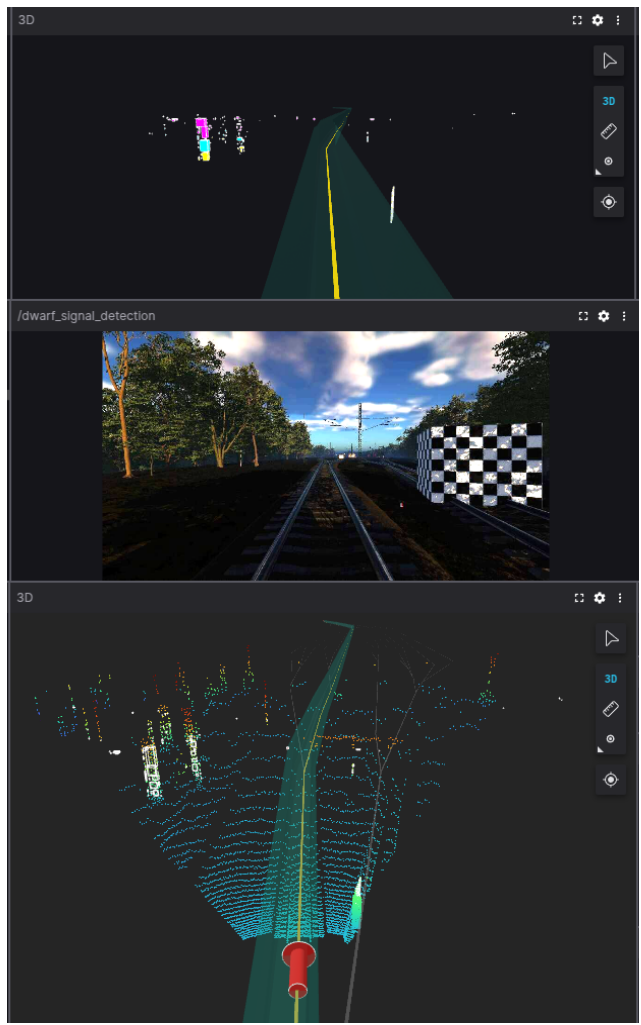

*Figure 6: ROS2 topic visualization of KaITrain (using foxglove studio), upper: clustered objects, ego track 112 (yellow) and clearance gauge (green); mid: RGB image; lower: 3d LiDAR scan, ego vehicle (red arrow), yard map (grey) ego Track(yellow)*

## VI. Conclusion and Outlook

This paper demonstrated how a successive ODD formulation could be established for the automation of a shunting locomotive, enabling the construction of specific test cases for a given use case. To this end, a graphical scenario description blueprint was introduced. A simple manoeuvring task was used to demonstrate how virtual testing can be conducted in a closed loop. The open-source Karlsruhe Intelligent Train (KaITrain) system was used as the SUT, with particular focus on signal and obstacle detection during troubleshooting. The shunting task was broken down into four µODDs, and the SUT was verified in fifty-six test cases. Significant deficiencies were identified in the KaITrain system. Following the findings presented here, these weak points can now be specifically addressed. These include the adjustment of parameters for point cloud-based object clustering and the formulation of topics for the observer. Additionally, to ensure error-free automated shunting, a solution must be found to address map deviation, particularly in switch areas and on curves. The next step will be to analyse the use of the camera image for track head detection. This would enable the track's course to be corrected, particularly within the vehicle's field of view, and the projected clearance gauge to be adapted to the track's actual course. It has been demonstrated that virtual environments can be utilised for development and testing purposes, yielding results quickly with minimal resource requirements. One possible objective is to develop diverse virtual tests to facilitate the authorisation process for new automated systems.

Using µODDs and generic volumes can also be an interesting approach in parallel automation initiatives, such as those in the agriculture, aerospace and road sectors. The bottom-up methodology provides a manageable strategic starting point for development and licensing, offering an overview of existing, working (sub-)functions and future developments. Using explicit obstacles during testing and development is comprehensible but lacks completeness. Therefore, unknown obstacles or objects that were not considered during the development phases may lead to errors in later detection or classification. Using a safety argumentation on generic volumes could be an approach to avoid the issue of completeness when dealing with licensing authorities. Furthermore, if generic bodies can represent the majority of obstacles, the testing process can be significantly accelerated. This would save resources and, consequently, money.

Current developments in result evaluation attempt to establish comparable metrics for statements. To this end, numerical values, frequencies and deviations from the formulated target results (expected values) must be collected and summarised systematically.

## Statements and Declarations

This research did not receive any specific grant from funding agencies in the public, commercial, or non-profit sectors. The virtual environment is still under development and continuously improved. The data generated and used for this study will be published on demand, if possible, as it is a live emulation and proceeding within to the closed -loop setup. The authors declare that they have no competing interests.

## Acknowledgment

This work forms part of the ongoing virtual testing (virTrack) and automated driving stack (KaITrain) research projects at the Institute of Vehicle System Technology (FAST) at the Karlsruhe Institute of Technology (KIT).

## AUTHORS

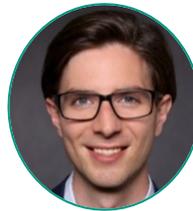

**Steffen Schäfer, M.Sc. RWTH,** He is a final-year PhD student and academic assistant at the Karlsruhe Institute of Technology, as well as research assistant at the Ohm, Nuremberg Institute of Technology. He leads a research team investigating virtual methods for functional verification and validation in the automation of rail vehicles.

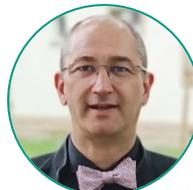

**Martin G. Cichon, Univ.-Prof. Dr.-Ing.** He is the Head of the Institute of Vehicle Systems Technology at the Karlsruhe Institute of Technology. His research focuses on complete vehicles, robust systems, electrical systems, and the application and approval readiness of highly automated systems.